\documentclass[conference]{IEEEtran}
\usepackage{comment}
\IEEEoverridecommandlockouts

\usepackage{cite}
\usepackage{amsmath,amssymb,amsfonts}
\usepackage{algorithmic}
\usepackage{graphicx}
\usepackage{textcomp}
\usepackage{makecell}
\usepackage{multirow}
\usepackage{orcidlink}
\usepackage{subfig}
\usepackage{xcolor}
\usepackage{url}
\usepackage{float}
\usepackage{hyperref}
\usepackage{gensymb}
\hypersetup{
    colorlinks=true,
    linkcolor=blue,
    filecolor=magenta,      
    urlcolor=blue,
    pdftitle={ForestProtector: An IoT Architecture Integrating Machine Vision and Deep Reinforcement Learning for Efficient Wildfire Monitoring},
    pdfauthor={Kenneth Bonilla-Ormachea; Horacio Cuizaga; Edwin Salcedo; Sebastian Castro; Sergio Fernandez-Testa; Misael Mamani},
  pdftitle={ForestProtector: An IoT Architecture Integrating Machine Vision and Deep Reinforcement Learning for Efficient Wildfire Monitoring},
  pdfsubject={Wildfire monitoring; IoT; computer vision; deep reinforcement learning},
  pdfkeywords={wildfire detection, IoT, computer vision, deep reinforcement learning, edge computing},
  pdfpagemode=UseNone,
    }
 
\def\BibTeX{{\rm B\kern-.05em{\sc i\kern-.025em b}\kern-.08em
    T\kern-.1667em\lower.7ex\hbox{E}\kern-.125emX}}
\begin{document}

\title{ForestProtector: An IoT Architecture Integrating Machine Vision and Deep Reinforcement Learning for Efficient Wildfire Monitoring
\thanks{This work was supported by a grant from Google AI, under the program TensorFlow Faculty Award 2021.}
}
  
\author{
\IEEEauthorblockN{
Kenneth Bonilla-Ormachea \orcidlink{0009-0004-1999-1630}, Horacio Cuizaga \orcidlink{0009-0002-7450-7632}, Edwin Salcedo \IEEEauthorrefmark{1}\orcidlink{0000-0001-8970-8838}, Sebastian Castro \orcidlink{0009-0002-0717-8031}, \\ 
Sergio Fernandez-Testa \orcidlink{0000-0003-3683-7497}, and Misael Mamani \orcidlink{0000-0003-1588-5625}
}
\IEEEauthorblockA{ 
Centro de Investigación, Desarrollo e Innovación en Ingeniería Mecatrónica\\
Universidad Católica Boliviana ``San Pablo'', La Paz, Bolivia \\
\IEEEauthorrefmark{1} Correspondence: esalcedo@ucb.edu.bo}
}

\maketitle

\begin{abstract}

Early detection of forest fires is crucial to minimizing the environmental and socioeconomic damage they cause. Indeed, a fire's duration directly correlates with the difficulty and cost of extinguishing it. For instance, a fire burning for 1 minute might require 1 liter of water to extinguish, while a 2-minute fire could demand 100 liters, and a 10-minute fire might necessitate 1,000 liters. On the other hand, existing fire detection systems based on novel technologies (e.g., remote sensing, PTZ cameras, UAVs) are often expensive and require human intervention, making continuous monitoring of large areas impractical. To address this challenge, this work proposes a low-cost forest fire detection system that utilizes a central gateway device with computer vision capabilities to monitor a 360° field of view for smoke at long distances. A deep reinforcement learning agent enhances surveillance by dynamically controlling the camera's orientation, leveraging real-time sensor data (smoke levels, ambient temperature, and humidity) from distributed IoT devices. This approach enables automated wildfire monitoring across expansive areas while reducing false positives. The full implementation, datasets, and trained models are available at \url{https://github.com/EdwinTSalcedo/ForestProtector}. 

\end{abstract}


\section{Introduction}

In recent years, a surge in forest fires has plagued the globe, fueled by a complex interplay of factors including climate change, deforestation, land use changes, human activities, and prolonged droughts. This escalating crisis has had a particularly devastating impact on Bolivia, where wildfires in 2024 alone have scorched over 10 million hectares (24.7 million acres) of land \cite{deutschewelle2024}. This staggering figure surpasses previous years, marking a significant environmental crisis for the country \cite{copernicus2024}. While developed nations utilize sophisticated early wildfire warning systems like ALERTCalifornia \cite{alertcalifornia2024}, NASA's FIRMS program \cite{nasa2024}, and the European Forest Fire Information System \cite{effis2024}, their replication in developing countries remains a challenge due to financial constraints. 

The longer a forest fire lasts, the more difficult and costly it becomes to extinguish. Therefore, early detection of forest fires is paramount in minimizing the environmental and socioeconomic damage they inflict, which in turn requires a low rate of false alarms. Fortunately, recent advances in embedded systems, artificial intelligence (AI), and the Internet of Things (IoT) are significantly enhancing the automation of wildfire monitoring and management. Novel real-time fire detection systems utilize a combination of sensors, microphones, and cameras to identify early fire indicators like smoke, heat, sound, and gas emissions \cite{dampage2022, benzekri2020, peruzzi2023, li2024}. 

While current technologies have significantly improved wildfire monitoring, there is a critical need for more accessible and efficient early detection methods. Consequently, this project explores a novel approach by integrating deep reinforcement learning (DRL), computer vision, and the Internet of Things (IoT) to enhance wildfire time-to-detection. Our proposed system gathers real-time sensor data (temperature, humidity, and smoke) from multiple locations. This data informs a DRL agent located within a central gateway. Then, the agent prioritizes the surveillance of areas to verify the presence of smoke at long distances using a 3D convolutional neural network (3DCNN). This monitoring strategy, supported by IoT and a Low-Power Wide-Area Network (LPWAN), offers an innovative approach for enhancing wildfire detection over vast areas and to reduce false positives.

\section{Background}
\label{sec:background}

\begin{figure*}
    \centering
    \includegraphics[width=\textwidth]{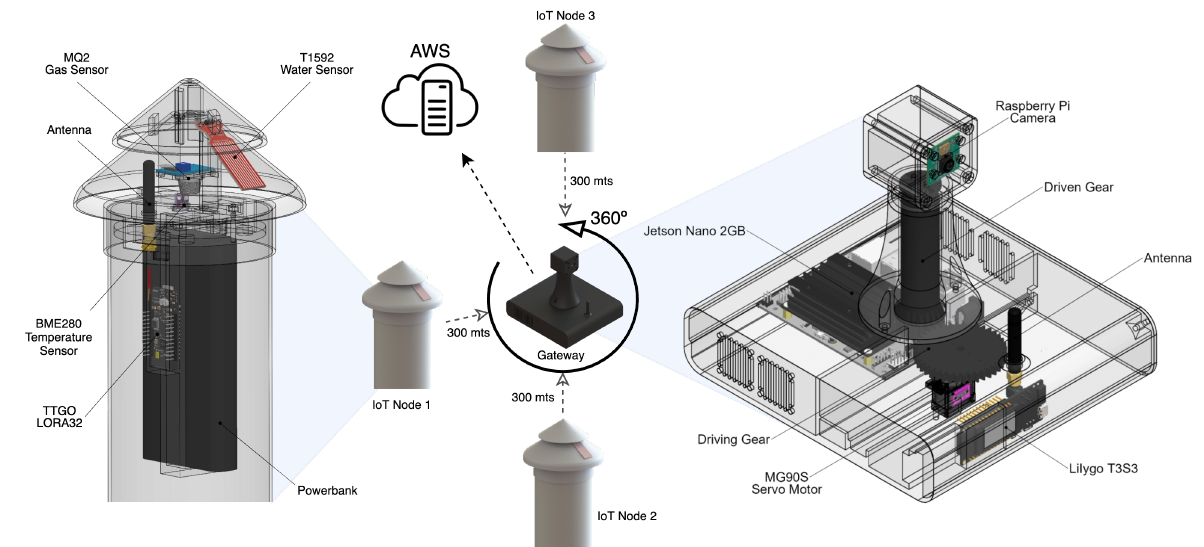} 
    \caption{Overview of the ForestProtector architecture for efficient wildfire monitoring.}
    \label{fig:system-architecture}
\end{figure*}

Conventional fire detection methods, often relying on human observation from watchtowers or using old tools such as the Osborne fire finder \cite{ororatech2024}, are inherently inefficient and prone to human error.   Over time, different wildfire detection technologies have emerged, with advantages and disadvantages involved. These solutions can be clustered into three main categories: remote sensing-based, optical sensor-based, and IoT sensor-based. Remote sensing (e.g. from Low Earth Orbit (LEO) satellites) is particularly valuable for monitoring large forest areas at a continental or global scale \cite{chen2024}. Although higher resolution and frequency in satellite imagery generally translate to increased costs, ongoing advancements in satellite technology and sensor development are improving revisit times and accessibility for wild fire detection \cite{yang2024}. This trend points towards greater accessibility and wider use of satellite-based fire detection in developing countries in the years to come.  

Unlike the optical sensors used in remote sensing, ground-based optical sensors are widely available and have been adopted for a wide range of applications due to their affordability and ease of use. These sensors can be deployed in diverse locations, including ground level, watchtowers, poles, and even unmanned aerial vehicles \cite{mukhiddinov2022, saydirasulovich2023}. Optical sensors fall into two main categories, RGB cameras and infrared (IR) cameras, with IR encompassing short-wave, middle-wave, and long-wave \cite{ciprian2023}. All of them ease the detection of fire traits, such as smoke or flames during day and night time. Furthermore, 360° cameras (also known as PTZ cameras) are frequently used \cite{shah2021}, despite their higher current cost.

Wireless sensor networks (WSNs), which can be integrated into the Internet of Things (IoT), can collect data on various factors, such as temperature, humidity, pressure, and gas levels (e.g., carbon monoxide and carbon dioxide). Indeed, some investigations have used this approach to continuously calculate fire risk \cite{dampage2022, benzekri2020}. However, as with optical-based sensing, WSNs face limitations in covering large areas due to the need for multiple nodes to achieve comprehensive monitoring \cite{lloret2009}. On the other hand, WSNs can be combined with other sensing modalities for prompt fire detection. For example, Peruzzi et al. \cite{peruzzi2023} embedded convolutional neural networks (CNNs) on low-power IoT devices for audio and image data classification in forest areas. The combined use of the CNNs provided higher accuracy (96.15\%) for normal or fire classification.

\section{Proposed System}
\label{sec:proposed-system}

The proposed system architecture, shown in Figure \ref{fig:system-architecture}, consists of two main components: IoT sensor nodes and a central gateway. The sensor nodes, deployed near forests, monitor environmental conditions using various sensors, including those for temperature, humidity, barometric pressure, smoke and water. The central gateway, based on a NVIDIA Jetson Nano card, collects data from these nodes via the LoRa protocol. The gateway then utilizes a DRL agent to control the gateway camera's perspective, focusing on potential nearby campfires, and employs a computer vision algorithm to verify the presence of smoke in the high-risk region. These components will be further described in the following sections. 

\begin{table}[H]
\caption{Sensors used in each IoT node.}
\begin{center}
\begin{tabular}{|c|c|c|c|}
\hline
\textbf{Sensor} & \textbf{Measurement} & \textbf{Range} & \textbf{Resolution} \\
\hline
\multirow{3}{*}{BME280} & Temperature & -40 °C to 85 °C & 0.01 °C \\
    & \makecell{Barometric\\ Pressure} & 300 hPa to 1100 hPa & 0.18 Pa \\
    & Humidity & 0\% to 100\% RH & 0.008\% RH \\
    \hline
   MQ2 & Smoke & 0 to 4095 & 1 \\
   \hline
   T1592 & Water & 0 a 4095 & 1 \\
\hline
\end{tabular}
\label{tab:sensors}
\end{center}
\end{table}

\subsection{IoT Sensor Nodes}

To collect environmental data, we developed three IoT sensor nodes. Each node consists of a BME280 sensor, an MQ2 sensor, a TTGO LoRa32 module, a T1592 sensor, and a power bank (See Table \ref{tab:sensors} for sensor specifications). The BME280 sensor measures temperature, humidity, and pressure, while the MQ2 sensor detects smoke particles in the air. T1592 sensors were included to detect precipitation and halt data collection during rain. The TTGO LoRa32 modules continuously transmit sensor data to the central gateway via the LoRaWAN protocol. To protect the IoT nodes in outdoor conditions (e.g., rain and solar radiation), we designed a custom case in SolidWorks and 3D printed it with PETG (Polyethylene Terephthalate Glycol) material. The design allows for mounting the node to the top of a 4-inch diameter PVC tube to be installed at ground-level.

\subsection{Central Gateway and Cloud Server}

The gateway uses a Lilygo T3S3 module to gather data from the IoT devices via the LoRaWAN protocol. Within the gateway, this data is transmitted in JSON format to the Jetson Nano, which acts as the central processing unit. A RL agent inside the Jetson Nano, trained on the IoT sensor data provided by the end devices, defines the Raspberry Picam v2 camera's horizontal orientation (ranging from 0\degree  to 360\degree). Once positioned, the camera, coupled with a computer vision system, detects smoke — an early indicator of wildfire. If smoke is detected, the gateway sends an alert to a cloud server, which then relays the alert to the designated users.

\begin{figure}[H]
    \centering
    \includegraphics[width=0.5\textwidth]{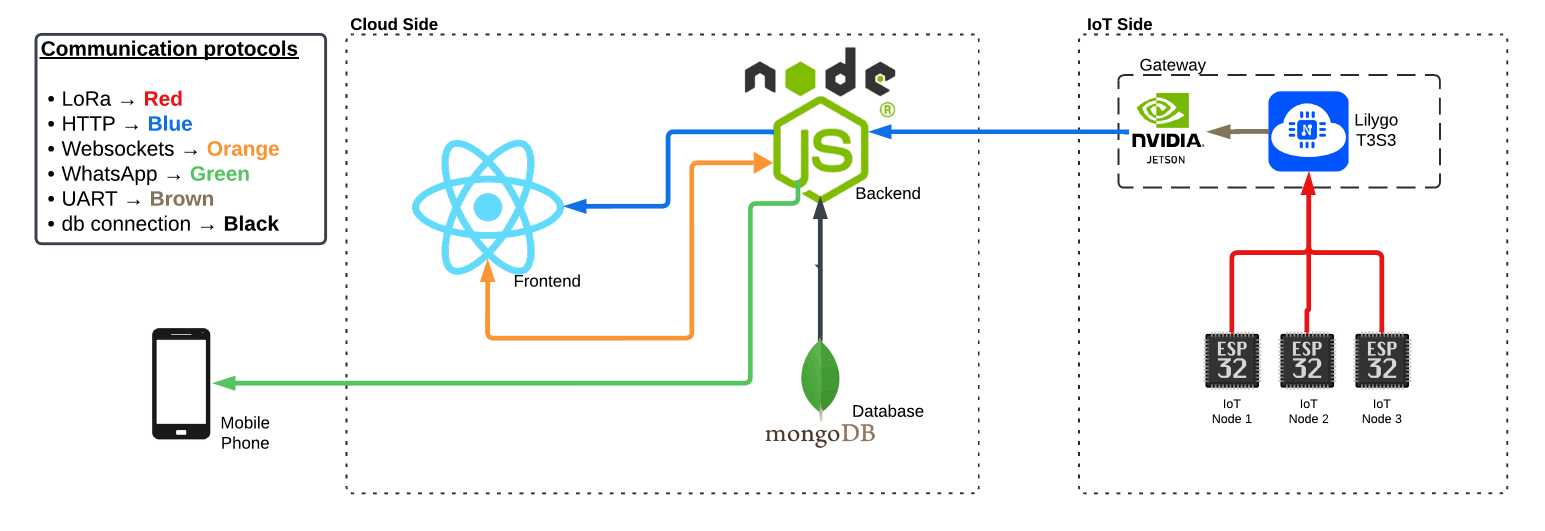}
    \caption{System's software architecture.}
    \label{fig:system-communication}
\end{figure}

Additionally, all sensor data are transmitted to the cloud, which is received by an Express.js backend and stored in a MongoDB database for analysis and monitoring. A user-friendly React.js frontend deployed on Vercel provides real-time data visualizations and creates push notifications using a socket connection. Moreover, the system architecture, deployed on an AWS EC2 instance, includes an alert system that sends WhatsApp notifications to a designated phone number using the WhatsAppWeb.js API and displays alerts on the dashboard. This system architecture is depicted in Figure \ref{fig:system-communication}. 

\section{Deep Reinforcement Learning Model}

\subsection{Model design and training}
The agent was designed using a Deep Q-Network (DQN) architecture, suitable for environments with discrete action spaces and complex observational patterns. The DQN model’s objective was to learn a policy that maximizes the cumulative reward by selecting the most relevant sector for monitoring based on indicators of potential wildfire risk. The resulting architecture consists of three fully connected layers: an input layer that processes the environment’s state, two hidden layers of 24 neurons with ReLU activations, and an output layer representing the Q-values for each action.

The Q-value function was updated using the Bellman equation, which models the expected cumulative reward for taking action $a$ in state $s$, followed by an optimal policy. The update rule is given by:

\begin{equation}
Q(s, a )= Q(s, a) + \alpha \Big( r + \gamma \max_{a'} Q(s', a') - Q(s, a) \Big)
\end{equation}

where:
$Q(s, a)$ is the estimated value of taking action $a$ in state $s$.
$\alpha$ is the learning rate, which controls the weight given to new information during the update.
$r$ is the reward received after taking action $a$ in state $s$.
$\gamma$ is the discount factor, which balances the importance of immediate versus future rewards.
$\max_{a'} Q(s', a')$ represents the maximum expected value of future rewards for the next state $s'$, considering all possible actions $a'$. 

\subsection{Decision-making process for detecting wildfires}

Each IoT sensor node provides observations on smoke levels, temperature, barometric pressure, humidity, and water presence, however, only temperature, humidity, and smoke levels were considered for the model. The model computes a weighted sum of these observations to prioritize the sector of the IoT node $i$ with the strongest indicators of potential wildfire presence. This signal strength for each IoT node $i$  is calculated as follows:

\begin{equation}
\text{Signal}_{\text{i}} = W_s * \frac{C_s}{100} + W_t * B_t + W_h * B_h
\label{eq:signal_strength}
\end{equation}

where $W_s$, $W_t$, and $W_h$ are the respective weights assigned to smoke, temperature, and humidity. $B_t$ and $B_h$ are binary values (1 or 0) indicating if thresholds for temperature and humidity have been surpassed. $C_s$ represents the smoke level, normalized to a scale from 0 to 1 by dividing by 100. 

Based on the calculated signal for each IoT node $i$, the agent selects the action $a^*$ that maximizes the expected reward according to Equation \ref{eq:optimal-action}. This approach allows the agent to focus on IoT nodes with the strongest evidence of wildfire risk, enhancing the model’s responsiveness to critical conditions.

\begin{equation}
a^* = \arg \max_{a} Q(s, a)
\label{eq:optimal-action}
\end{equation}

Moving on, the performance of the DQN agent was evaluated through two primary metrics: the moving average of rewards over episodes and the training loss over time. Together, these metrics validate the DQN agent's ability to effectively learn and optimize its policy within the given environment. The moving average, calculated over a 100-episode window, provides a smooth indicator of the agent's learning progress. This is formally defined in Equation \ref{eq:rl-first_metric}, for a given episode \( t \), the moving average of rewards \( \overline{R}_t \) over a window \( w \). In this equation, \( R_k \) denotes the total reward received in episode \( k \).

\begin{equation}
\overline{R}_t = \frac{1}{w} \sum_{k=t-w+1}^{t} R_k
\label{eq:rl-first_metric}
\end{equation}

As for the Q-network, the training loss provides insights into the convergence. It is computed as the mean squared error (MSE) between the predicted Q-value and the target Q-value for each state-action pair in a batch. Given a mini-batch of size \( N \), the MSE loss \( L \) can be defined as:

\begin{equation}
L = \frac{1}{N} \sum_{i=1}^{N} \left( Q(s_i, a_i) - \left[ r_i + \gamma \max_{a'} Q(s_{i+1}, a') \right] \right)^2
\label{eq:rl-second_metric}
\end{equation}

\begin{figure*}
    \centering
    \includegraphics[width=\textwidth]{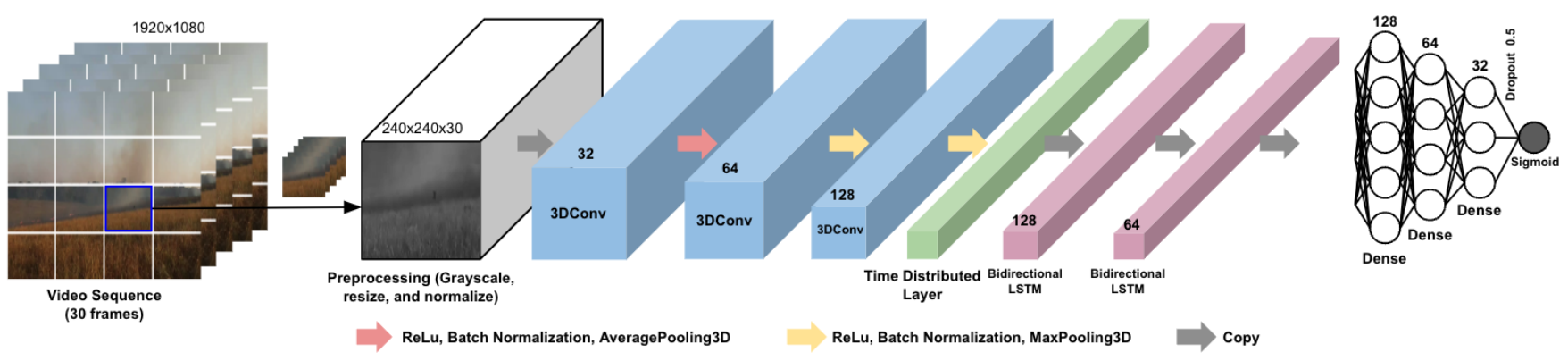} 
    \caption{S5, the best performing 3DCNN model for smoke detection at long distances.}
    \label{fig:3dcnn}
\end{figure*}

\section{Wildfire Detection Using Computer Vision}

The proposed computer vision system utilizes a Raspberry Pi Camera Module v2.1 installed in the gateway, enabling the collection of RGB images with a resolution of 1920px x 1080px at 30 frames per second. During the day, the system prioritizes smoke detection over fire detection, as smoke is a more visible wildfire indicator from long distances. Inspired by our successful results classifying video on embedded devices (Jetson Nano development cards) \cite{fernandez2024}, we implemented 3DCNN models for smoke recognition. These models were trained using a combination of real and synthetic videos of smoke plumes and wildfire captured at various distances. 

For nighttime fire detection, the system relies on a modified version of the algorithm proposed by Gunay et al. in \cite{gunay2009}. Our implementation prioritizes bright region detection (BRD), average magnitude difference function (AMDF), background subtraction, and color thresholding. We collected 50 night videos with fire and 50 night videos without fire. Through experimentation, we identified that 180 and 0.2 were suitable threshold values for the BRD and AMDF components, respectively. In the following sections, we describe the development of the 3DCNN-based model for daytime wildfire detection.

\subsection{Dataset preparation}

To collect the dataset detailed in Table \ref{tab:video-dataset}, we used video editing tools, scraped videos from social media websites, and used generative AI tools. First, we used After Effects to generate smoke effects with varying wind and orientation, which were then fused with videos of forests recorded in La Paz, Bolivia. We also scraped videos from social media using search terms like ``wildfire,'' ``burning forest,'' and ``wild forest.'' Finally, we used the AI tool Haiper \cite{haiperai2024} to generate wildfire videos from static images.  

We used data augmentation to create the final dataset, applying the following transformations: Horizontal Flip, Random Brightness and Contrast, Gaussian Blur, and Gaussian Noise. Each transformation had a 50\% probability of being applied sequentially using the Albumentations library. This process resulted in a final dataset of 4,000 videos. For experimentation purposes, all videos were converted to grayscale, cropped to a 1:1 aspect ratio, and resized to 240x240 pixels.

\begin{table}[H]
\caption{Data collection strategies applied to train the smoke detection model. The last column denotes the final subsets after applying data augmentation. }
\begin{center}
\begin{tabular}{|c|c|c|c|c|c|}
\hline
\textbf{Class} & \textbf{\makecell{Scrapped \\Videos}} & \textbf{\makecell{AI-\\generated \\Videos}} & \textbf{\makecell{Videos\\ Generated \\with \\After \\Effects}} & \textbf{Total} & \textbf{Aug.} \\
\hline
\makecell{With \\Smoke}    & 108                      & 443               & 184                & 735    & 2,000        \\ \hline
\makecell{Without \\Smoke}   & 663                      & 0                 & 0                  & 663    & 2,000          \\ \hline
\textbf{Total}   & 771                      & 443                 & 184                  & 1,398    & 4,000          \\ \hline
\end{tabular}
\label{tab:video-dataset}
\end{center}
\end{table}

\subsection{3DCNN Architecture}
This study developed and evaluated five distinct 3DCNN architectures (S1 to S5) to optimize smoke detection accuracy in video sequences. The first model, S1, used Conv3D layers with MaxPooling3D to capture spatial and temporal features, followed by dense layers for classification. S2 built upon S1 by replacing MaxPooling3D with AveragePooling3D, which smoothed feature extraction and improved accuracy. S3, which maintained the structure of S2 but used grayscale data, achieved the best results during testing across all architectures, reducing computational load in the process. The conversion to grayscale was performed after observing the superior performance of models trained on grayscale data. S4 extended S2 with BatchNormalization to stabilize training and TimeDistributed layers to prepare data for bidirectional LSTM layers, further enhancing temporal feature learning. Finally, S5 replicated S4 but utilized grayscale data. Figure \ref{fig:3dcnn} illustrates the architecture of S5, the best-performing model.

\section{Experimental Results and Discussion}
\label{sec:results}

\subsection{Deep Reinforcement Learning Model}

\begin{figure}[H]
    \centering
    \includegraphics[width=0.48\textwidth]{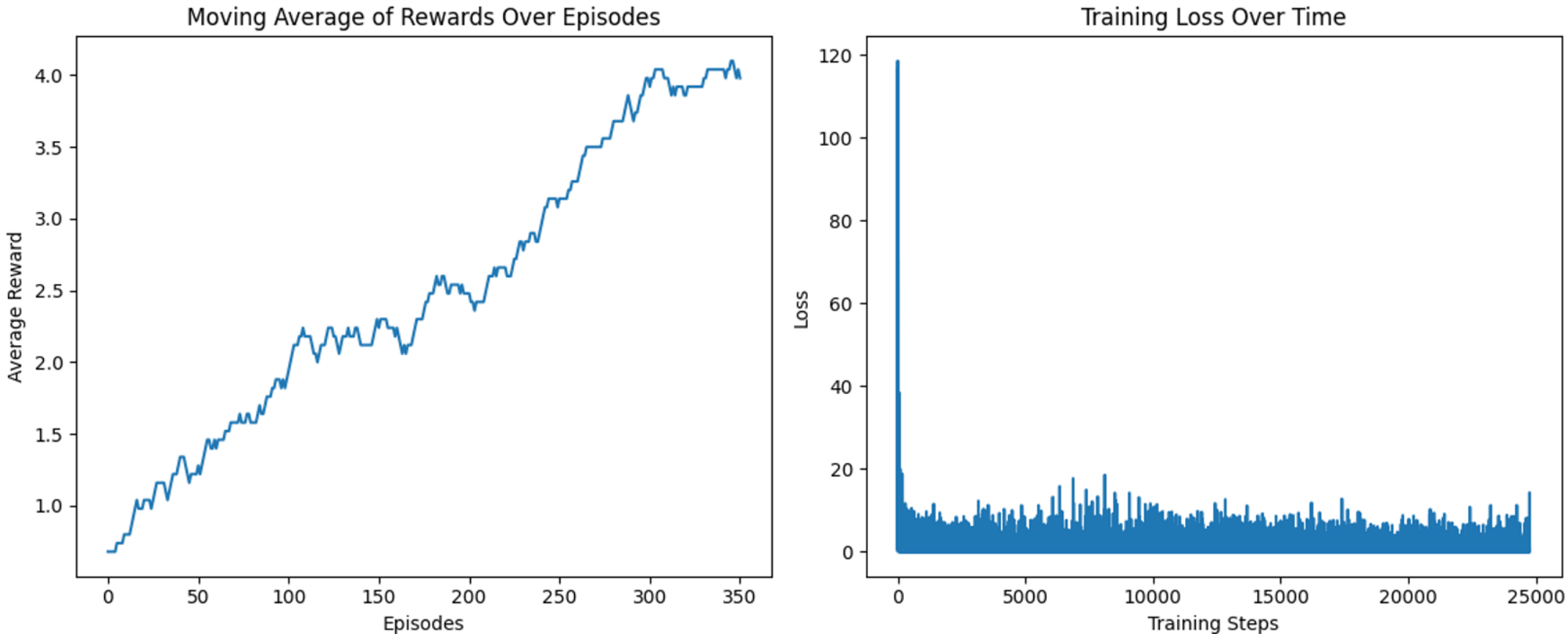}
    \caption{Moving Average of Rewards and Training Loss over Time, showing the agent's learning progress and convergence.}
    \label{fig:results_Rf}
\end{figure}

We trained the model using the following hyperparameters: learning rate $\alpha = 0.001$, discount factor $\gamma = 0.75$ (to balance the importance of immediate and future rewards), and rewards $r = 5$ for correctly identifying the IoT node with the strongest signal, and $r = -1$ otherwise. To enhance stability during training, an experience replay buffer of size 100,000 was employed to store past transitions. At each step, a mini-batch of 64 transitions was randomly sampled. The epsilon-greedy policy enabled a balance between exploration and exploitation, with $\epsilon$ initialized at 1.0 and decaying at a rate of 0.995 per episode, down to a minimum of 0.01. As illustrated in Figure~\ref{fig:results_Rf}, the model demonstrated significant improvement in its decision-making capability over the episodes. Additionally, the following weights were assigned to the sensor data: $W_s = 0.6$ (highest priority, as it provides a continuous measure strongly correlated with wildfire risk), $W_t = 0.3$ (temperature sensor contributes significantly when combined with humidity), and $W_h = 0.05$ (humidity sensor has the lowest individual weight). 

\subsection{Smoke Detection}

\begin{figure}[H]
    \centering
    \includegraphics[width=0.5\textwidth]{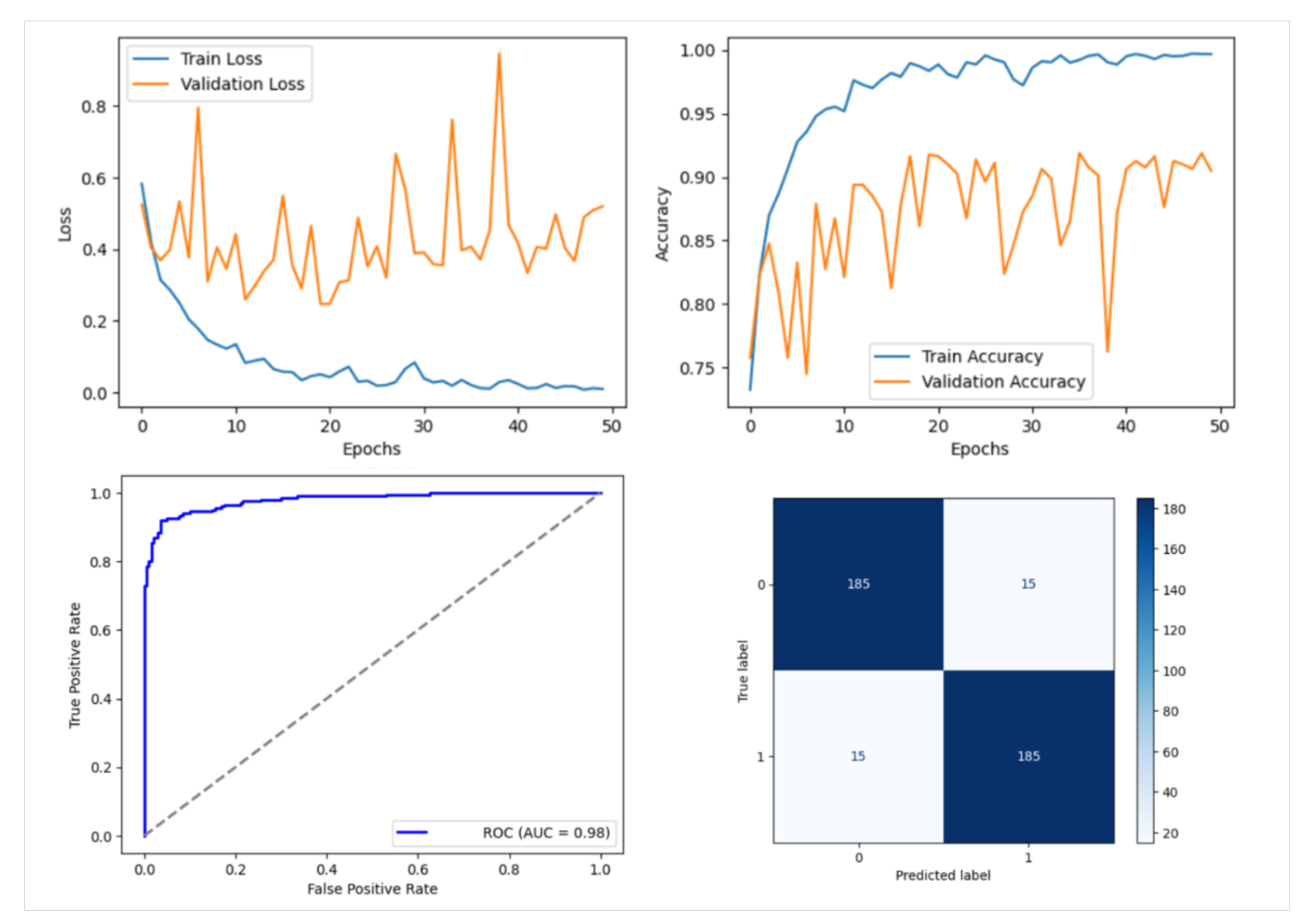}
    \caption{Performance metrics of the S5 model, including loss, accuracy, AUC, and confusion matrix.}
    \label{fig:train-metrics}
\end{figure}

\begin{figure}[H]
    \centering
    \includegraphics[width=0.5\textwidth]{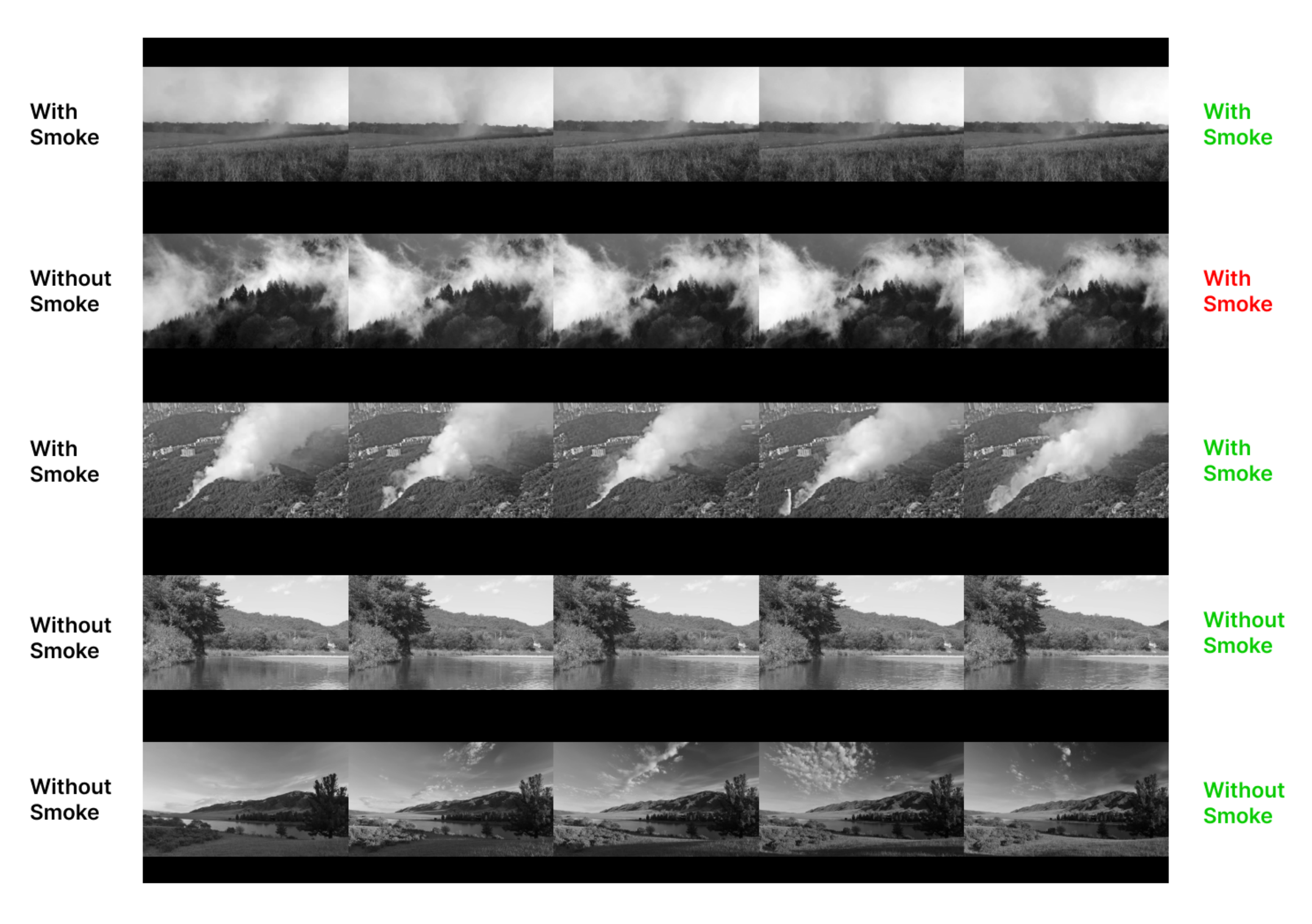}
    \caption{Correct classification and misclassification examples. True labels are on the left and predicted ones on the right. }
    \label{fig:3dcnn-samples}
\end{figure}

Table \ref{tab:comparison} provides a comparative analysis of the five 3DCNN architectures (S1 to S5) experimented in the project. Our experiments demonstrated the importance of incorporating modules like Average Pooling, bidirectional LSTM layers, and grayscale conversion for enhanced performance. S5 achieved the highest metrics (accuracy = 0.925, F-score = 0.93), demonstrating superior performance. Figure \ref{fig:train-metrics} shows the accuracy and loss over epochs, highlighting S5's convergence and stability. Furthermore, Figure \ref{fig:3dcnn-samples} shows classification examples obtained using S5, where it is worth noting that the model misclassified fog as smoke in the second sample. In the inference stage, the video captured by the camera is divided into an 8x4 grid, and each cell is analyzed individually. 

\begin{table}[H]
\caption{Comparative analysis of 3DCNN models for smoke detection.}
\begin{center}
\begin{tabular}{|c|c|c|c|c|}
\hline
\textbf{Model} & \textbf{AUC} & \textbf{Recall} & \textbf{F1-score} & \textbf{Accuracy} \\
\hline
S1 & 0.93 & 0.87 & 0.87 & 0.83 \\
S2 & 0.95 & 0.89 & 0.89 & 0.89 \\
S3 & 0.95 & 0.90 & 0.89 & 0.89 \\
S4 & 0.96 & 0.89 & 0.89 & 0.89 \\
S5 & 0.98 & 0.93 & 0.93 & 0.925 \\
\hline
\end{tabular}
\label{tab:comparison}
\end{center}
\end{table}

\subsection{System Performance Results}

\begin{figure}[ht]
  \centering
  \centering
  \subfloat[IoT sensor node]{\includegraphics[width=0.24\textwidth]{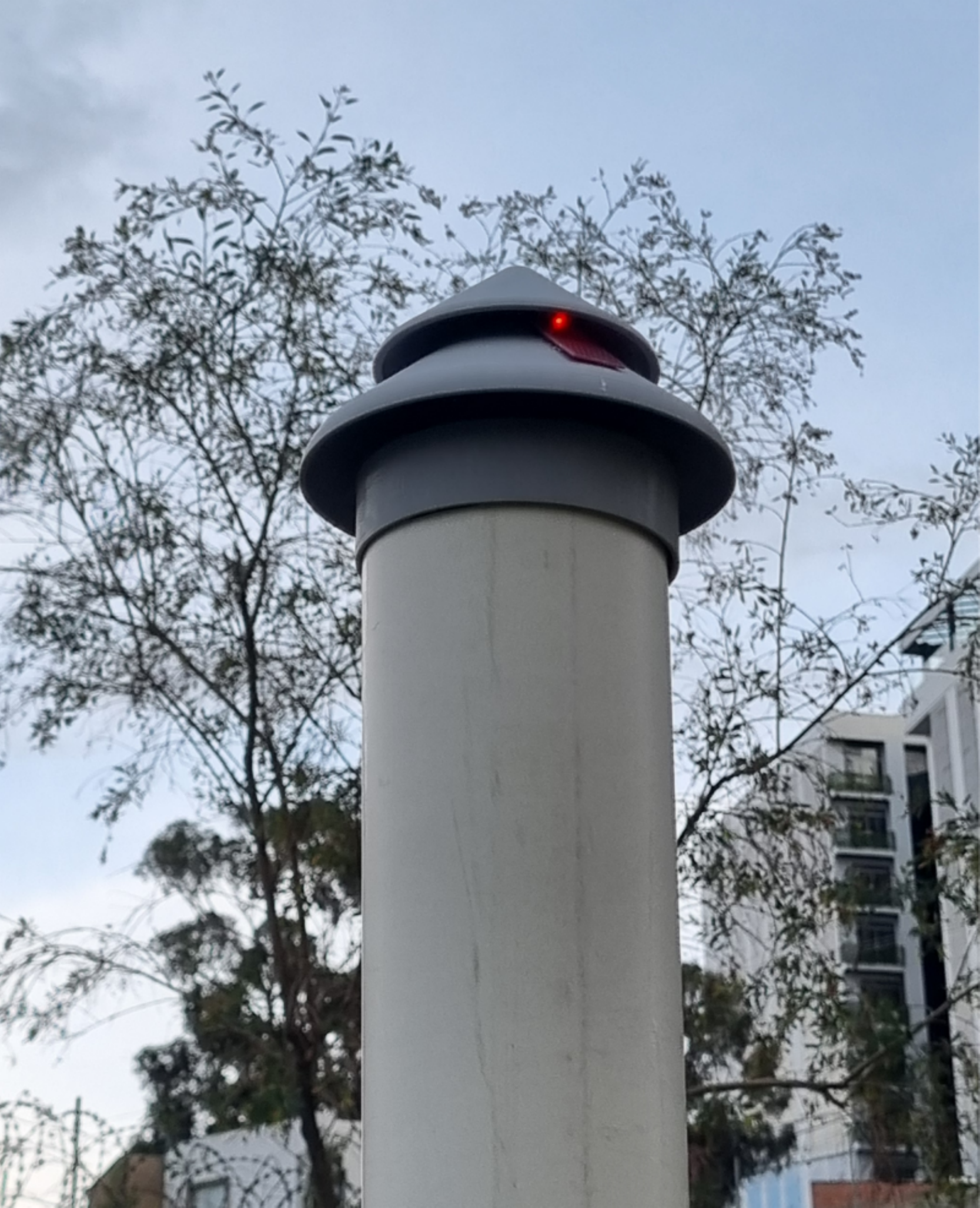}\label{fig:iot-device-prototype}}
  \hfill
  \subfloat[Gateway]{\includegraphics[width=0.24\textwidth]{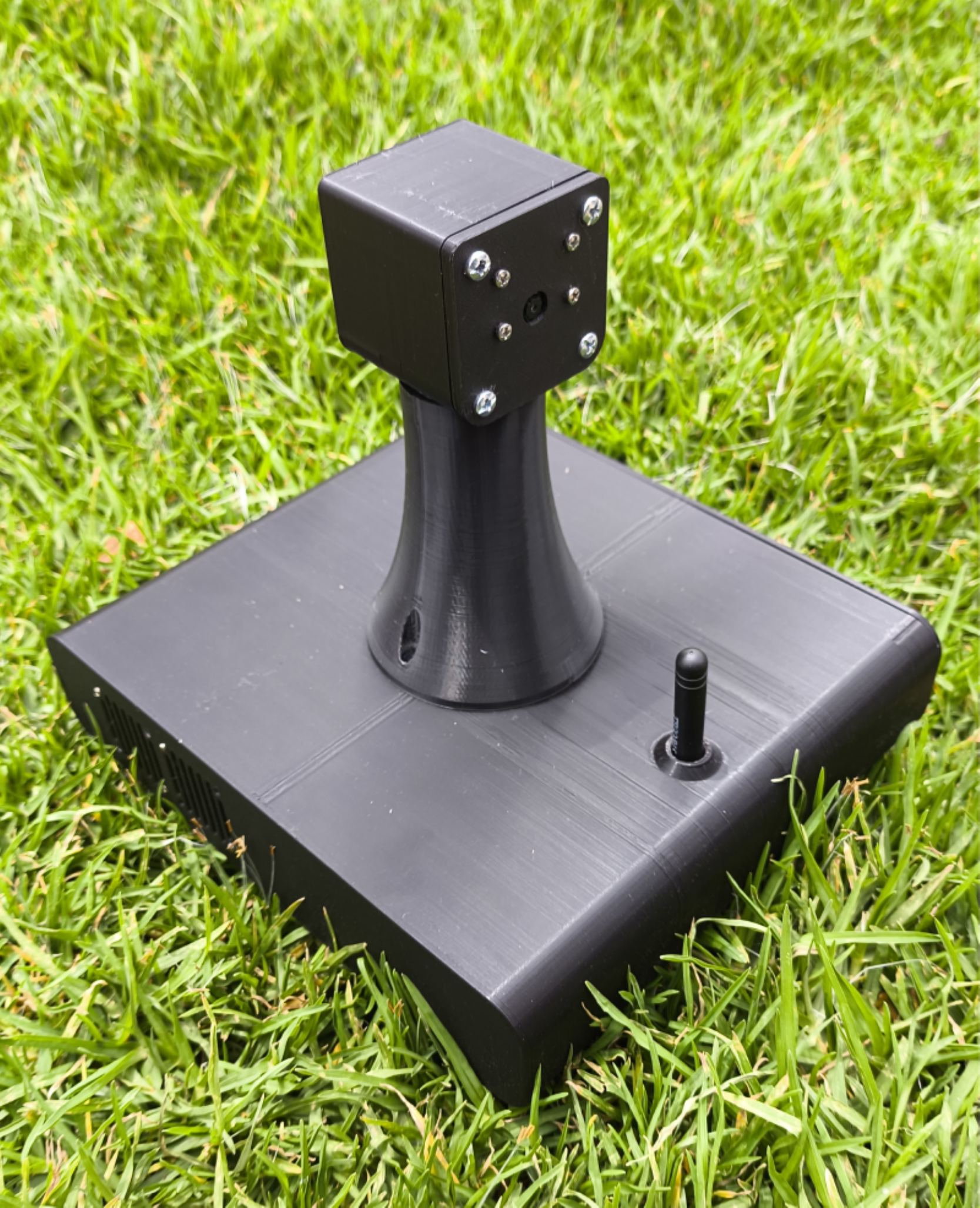}\label{fig:gateway-prototype}}

  \hfill
  \subfloat[Field testing setup]{\includegraphics[width=0.49\textwidth]{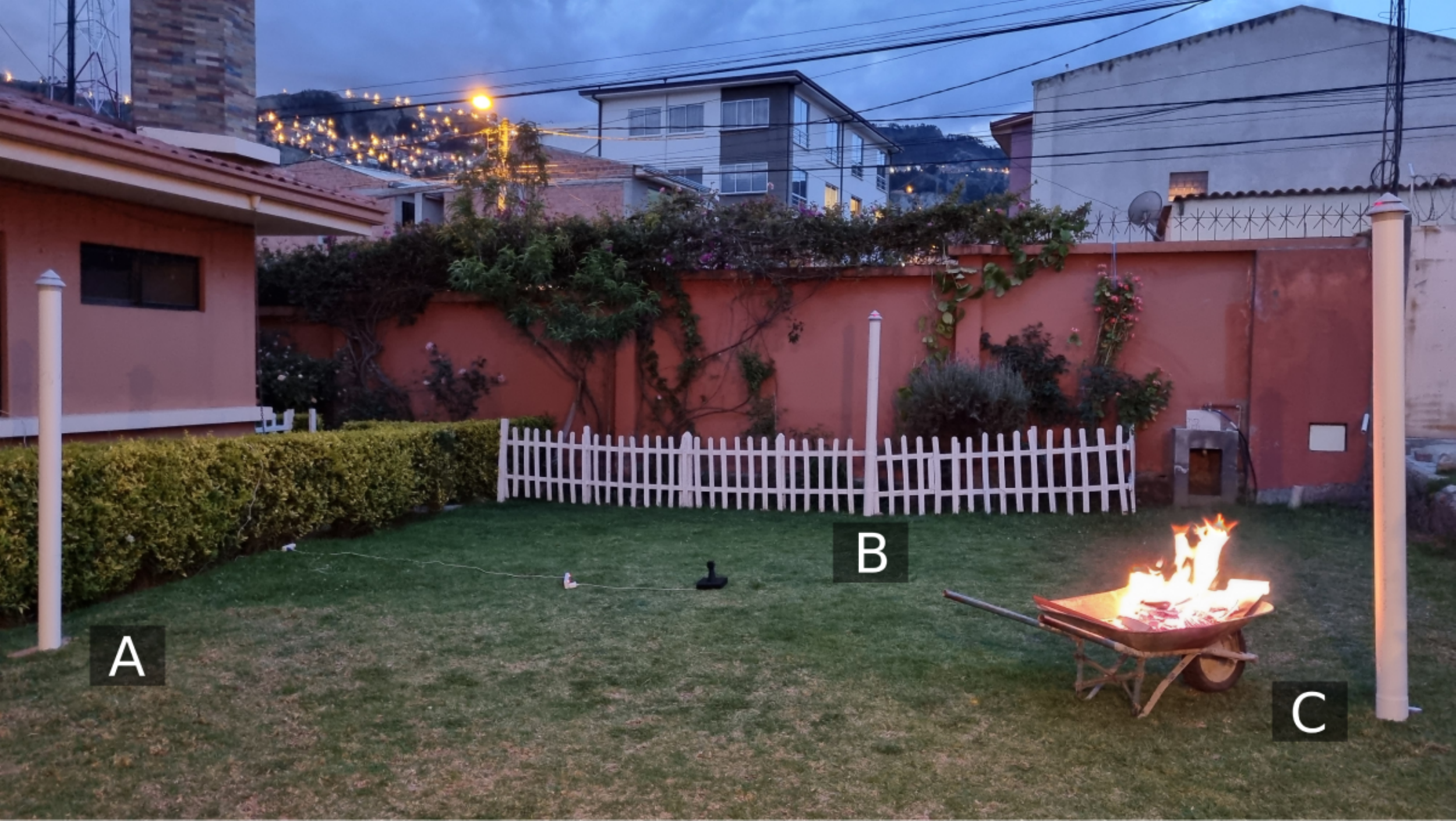}\label{fig:testing-setup}}
  
  \caption{Deployment of the system components for testing the fire detection approach.}
  \label{fig:final-prototype}
\end{figure}

The final prototypes of the IoT device and gateway are shown in Figures \ref{fig:iot-device-prototype} and \ref{fig:gateway-prototype}, respectively. On November 30th, 2024, we conducted a series of outdoor experiments in La Paz, Bolivia, to test the system's effectiveness in detecting fire features in real scenarios. As shown in Figure \ref{fig:testing-setup}, we installed the gateway in the middle and the IoT nodes at a distance of 3 meters from the gateway. We ignited a fire on a wheelbarrow. To simulate wildfire proximity, we moved the wheelbarrow closer to the IoT sensor nodes, keeping the fire facing outwards. We measured the time between moving the wheelbarrow closer to each node and the web application issuing an alert. These tests are shown in Table \ref{tab:experimental-tests}. The last column of this table shows the average detection time per IoT node induced by the fire on the wheelbarrow.

\begin{table}[H]
\caption{Experimental tests measuring the time in seconds taken by the system to detect and verify a fire event until sending an alarm to the designated users.}
\begin{center}
\begin{tabular}{|c|c|c|c|c|c|c|}
\hline
\textbf{\makecell{IoT \\node}} & \textbf{$t_1$} & \textbf{$t_2$} & \textbf{$t_3$} & \textbf{$t_4$} & \textbf{$t_5$} & \textbf{Mean} \\
\hline
A &118 &117 &118 &120 &117 &118 \\
\hline
B &124 &120 &123 &123 &122 &122.4 \\
\hline
C &129 &131 &128 &131 &130 &129.8 \\
\hline
\end{tabular}
\label{tab:experimental-tests}
\end{center}
\end{table}

Our experimental results in open environments demonstrate the effectiveness of the proposed system in detecting fires and generating timely alerts. The DRL agent effectively directed the camera towards the source of the fire, and the 3DCNN model accurately verified the presence of smoke. The DRL agent quickly defined the camera orientation towards the potential fire source in approximately 0.82 seconds. However, the 3DCNN required significantly longer—over 1.7 minutes on average—to process the multiple frame sequences of 224x224 dimensions, highlighting the potential for optimizing the 3DCNN model's inference time.

While our system showed promising results, its field deployment revealed several areas for improvement in future research. Firstly, using professional sensors for detecting smoke, CO, CO2, and other fire-related gases is crucial. While MQ2 sensors can detect various gases, including H2, LPG, CH4, CO, alcohol, smoke, and propane, their sensing range is limited. Also, they exhibited high sensitivity only for gases in close proximity to the sensor. Secondly, exploring the integration of solar energy could enhance the system's autonomy. Finally, further validation requires deploying and evaluating the system in diverse real-world environments and at varying distances. 

\section{Conclusions}
The proposed ForestProtector architecture integrates IoT technology, deep reinforcement learning, and computer vision to monitor large forest areas in real time. By leveraging IoT sensor nodes, a central gateway with a deep reinforcement learning agent, and a 3DCNN model for smoke detection, the system ensures effective risk prioritization and accurate smoke classification. Its modular design and cost-effective components make it scalable and accessible, offering a practical solution for wildfire detection in resource-constrained regions.

\label{sec:conclusions}

\bibliographystyle{ieeetr}
\bibliography{bibliography}

@article{yang2024,
  title={Advancements in remote sensing for active fire detection: a review of datasets and methods},
  author={Yang, Songxi and Huang, Qunying and Yu, Manzhu},
  journal={Science of the total environment},
  pages={173273},
  year={2024},
  publisher={Elsevier}
}

@article{shah2021,
  title={Preliminary Wildfire Detection Using State-of-the-art PTZ (Pan, Tilt, Zoom) Camera Technology and Convolutional Neural Networks},
  author={Shah, Samarth},
  journal={arXiv preprint arXiv:2109.05083},
  year={2021}
}

@misc{alertcalifornia2024,
    title = {ALERTCalifornia},
    author = {University of California},
    howpublished = {ALERTCALIFORNIA.org},
    url = {https://alertcalifornia.org/},
    note = {Accessed: May 2, 2024}
    }

@misc{deutschewelle2024,
    title = {How Land Clearance is Destroying Bolivia’s Forests},
    author = {Deutsche Welle},
    howpublished = {DW.com},
    url = {https://www.dw.com/en/behind-the-smokescreen-how-land-clearance-is-destroying-bolivias-forests/a-70466451},
    note = {Accessed: Oct 23, 2024}
    }

@misc{nasa2024,
    title = {Fire Information for Resource Management System},
    author = {NASA},
    howpublished = {NASA.gov},
    url = {https://firms.modaps.eosdis.nasa.gov/},
    note = {Accessed: Oct 2, 2024}
    }

@misc{effis2024,
    title = {European Forest Fire Information System EFFIS},
    author = {European Commision},
    howpublished = {COPERNICUS.eu},
    url = {https://forest-fire.emergency.copernicus.eu/},
    note = {Accessed: Oct 2, 2024}
    }

@misc{copernicus2024,
    title = {2023: A year of intense global wildfire activity},
    author = {European Commision},
    howpublished = {COPERNICUS.eu},
    url = {https://atmosphere.copernicus.eu/2023-year-intense-global-wildfire-activity},
    note = {Accessed: Oct 2, 2024}
    }

@article{dampage2022,
  title={Forest fire detection system using wireless sensor networks and machine learning},
  author={Dampage, Udaya and Bandaranayake, Lumini and Wanasinghe, Ridma and Kottahachchi, Kishanga and Jayasanka, Bathiya},
  journal={Scientific reports},
  volume={12},
  number={1},
  pages={46},
  year={2022},
  publisher={Nature Publishing Group UK London}
}

@article{benzekri2020,
  title={Early forest fire detection system using wireless sensor network and deep learning},
  author={Benzekri, Wiame and El Moussati, Ali and Moussaoui, Omar and Berrajaa, Mohammed},
  journal={International Journal of Advanced Computer Science and Applications},
  volume={11},
  number={5},
  year={2020},
  publisher={Science and Information (SAI) Organization Limited}
}

@article{peruzzi2023,
  title={Fight fire with fire: Detecting forest fires with embedded machine learning models dealing with audio and images on low power IoT devices},
  author={Peruzzi, Giacomo and Pozzebon, Alessandro and Van Der Meer, Mattia},
  journal={Sensors},
  volume={23},
  number={2},
  pages={783},
  year={2023},
  publisher={MDPI}
}

@article{chen2024,
  title={Remote sensing for wildfire monitoring: Insights into burned area, emissions, and fire dynamics},
  author={Chen, Yang and Morton, Douglas C and Randerson, James T},
  journal={One Earth},
  volume={7},
  number={6},
  pages={1022--1028},
  year={2024},
  publisher={Elsevier}
}

@article{mukhiddinov2022,
  title={A wildfire smoke detection system using unmanned aerial vehicle images based on the optimized YOLOv5},
  author={Mukhiddinov, Mukhriddin and Abdusalomov, Akmalbek Bobomirzaevich and Cho, Jinsoo},
  journal={Sensors},
  volume={22},
  number={23},
  pages={9384},
  year={2022},
  publisher={MDPI}
}

@article{saydirasulovich2023,
  title={An improved wildfire smoke detection based on YOLOv8 and UAV images},
  author={Saydirasulovich, Saydirasulov Norkobil and Mukhiddinov, Mukhriddin and Djuraev, Oybek and Abdusalomov, Akmalbek and Cho, Young-Im},
  journal={Sensors},
  volume={23},
  number={20},
  pages={8374},
  year={2023},
  publisher={MDPI}
}

@misc{ororatech2024,
    title = {Why Earth Observation is the Future of Fire Detection},
    author = {Orora Tech},
    howpublished = {ORORATECH.com},
    url = {https://ororatech.com/why-earth-observation-is-the-future-of-fire-detection/},
    note = {Accessed: Oct 30, 2024}
    }

@article{ciprian2023,
  title={FIRe-GAN: a novel deep learning-based infrared-visible fusion method for wildfire imagery},
  author={Cipri{\'a}n-S{\'a}nchez, Jorge Francisco and Ochoa-Ruiz, Gilberto and Gonzalez-Mendoza, Miguel and Rossi, Lucile},
  journal={Neural Computing and Applications},
  pages={1--13},
  year={2023},
  publisher={Springer}
}

@article{lloret2009,
  title={A wireless sensor network deployment for rural and forest fire detection and verification},
  author={Lloret, Jaime and Garcia, Miguel and Bri, Diana and Sendra, Sandra},
  journal={sensors},
  volume={9},
  number={11},
  pages={8722--8747},
  year={2009},
  publisher={Molecular Diversity Preservation International}
}

@article{li2024,
  title={A Lightweight Convolutional Spiking Neural Network for Fires Detection Based on Acoustics},
  author={Li, Xiaohuan and Liu, Yi and Zheng, Libo and Zhang, Wenqiong},
  journal={Electronics},
  volume={13},
  number={15},
  pages={2948},
  year={2024},
  publisher={MDPI}
}

@article{gunay2009,
  title={Video based wildfire detection at night},
  author={G{\"u}nay, Osman and Ta{\c{s}}demir, Kas{\i}m and T{\"o}reyin, B U{\u{g}}ur and {\c{C}}etin, A Enis},
  journal={Fire Safety Journal},
  volume={44},
  number={6},
  pages={860--868},
  year={2009},
  publisher={Elsevier}
}

@misc{haiperai2024,
    title = {HAIPER},
    author = {HAIPER AI},
    howpublished = {HAIPER.ai},
    url = {https://haiper.ai/home},
    note = {Accessed: Nov. 1, 2024}
    }

@inproceedings{fernandez2024,
  author={Fernandez-Testa, Sergio and Salcedo, Edwin},
  booktitle={2024 37th SIBGRAPI Conference on Graphics, Patterns and Images (SIBGRAPI)}, 
  title={Distributed Intelligent Video Surveillance for Early Armed Robbery Detection Based on Deep Learning}, 
  year={2024},
  volume={},
  number={},
  pages={1-6},
  doi={10.1109/SIBGRAPI62404.2024.10716299}
}

\end{document}